\documentclass[10pt,twocolumn,letterpaper]{article}

\usepackage{iccv}
\usepackage{times}
\usepackage{epsfig}
\usepackage{graphicx}
\usepackage{amsmath}
\usepackage{amssymb}

\usepackage{multirow}
\usepackage{booktabs}
\usepackage[dvipsnames,table]{xcolor}
\usepackage{subfig}
\usepackage{bm}
\usepackage[accsupp]{axessibility} 

\usepackage[breaklinks=true,bookmarks=false]{hyperref}

\def\wrt{\mbox{\textit{w.r.t.~}}}

\iccvfinalcopy 

\def\iccvPaperID{3838} 

\ificcvfinal\fi

\makeatletter
\def\blfootnote{\xdef\@thefnmark{}\@footnotetext}
\makeatother

\begin{document}

\title{Prompt Switch: Efficient CLIP Adaptation for Text-Video Retrieval}

\author{Chaorui Deng$^{1,*}$, ~Qi Chen$^{1,*}$, ~Pengda Qin$^2$, ~Da Chen$^3$, ~Qi Wu$^{1,\dagger}$ \\
$^1$Australia Institute of Machine Learning, University of Adelaide\\
$^2$Alibaba Group, $^3$Department of Computer Science, University of Bath\\
}

\maketitle
\ificcvfinal\thispagestyle{empty}\fi

\begin{abstract}
In text-video retrieval, recent works have benefited from the powerful learning capabilities of pre-trained text-image foundation models (e.g., CLIP) by adapting them to the video domain. 
A critical problem for them is how to effectively capture the rich semantics inside the video using the image encoder of CLIP.
To tackle this, state-of-the-art methods adopt complex cross-modal modeling techniques to fuse the text information into video frame representations,
which, however, incurs severe efficiency issues in large-scale retrieval systems as the video representations must be recomputed online for every text query.
In this paper, we discard this problematic cross-modal fusion process and aim to learn semantically-enhanced representations purely from the video,
so that the video representations can be computed offline and reused for different texts.
Concretely, we first introduce a spatial-temporal ``{Prompt Cube}'' into the CLIP image encoder and iteratively switch it within the encoder layers to efficiently incorporate the global video semantics into frame representations.
We then propose to apply an auxiliary video captioning objective to train the frame representations, 
which facilitates the learning of detailed video semantics by providing fine-grained guidance in the semantic space.
With a naive temporal fusion strategy (i.e., mean-pooling) on the enhanced frame representations, we obtain state-of-the-art performances on three benchmark datasets, i.e., MSR-VTT, MSVD, and LSMDC.
\blfootnote{$^*$Co-first author. $^\dagger$Corresponds to~\texttt{qi.wu01@adelaide.edu.au}. Code: \texttt{https://github.com/bladewaltz1/PromptSwitch}.}
\end{abstract}

\section{Introduction}\label{sec:introduction}

Text-video retrieval~\cite{bain2021frozen,chen2020fine,miech2019howto100m,wang2021t2vlad} is a fundamental task in the area of video-language understanding that seeks to find the most relevant video from a large set of candidates to match a text query. 
With the rapid growth of video data, text-video retrieval has become increasingly important for various applications, including video recommendation~\cite{sang2020context,wei2019mmgcn}, video search~\cite{kratochvil2020somhunter,wu2020interpretable}, and video summarization~\cite{deng2021sketch,messaoud2021deepqamvs,narasimhan2021clip,plummer2017enhancing,saquil2021multiple,xiao2020query}. 
Due to the high cost of constructing text-video datasets, one promising approach for this task is to leverage pre-trained text-image foundation models and transfer their powerful representation capabilities to the video domain.
Specifically, the CLIP~\cite{radford2021learning} model, which is trained using a text-image alignment objective, is particularly suitable for text-video retrieval and has been frequently studied recently~\cite{luo2022clip4clip,wu2022cap4video,gorti2022x}.
It has two transformer encoders~\cite{vaswani2017attention,dosovitskiyimage} to process images and texts, respectively. 
A vector representation is extracted from the input of each modality with the corresponding encoder and is optimized to be close to its paired representation and away from the unpaired ones.

\begin{figure}[t]
    \centering
    \includegraphics[width=0.95\linewidth]{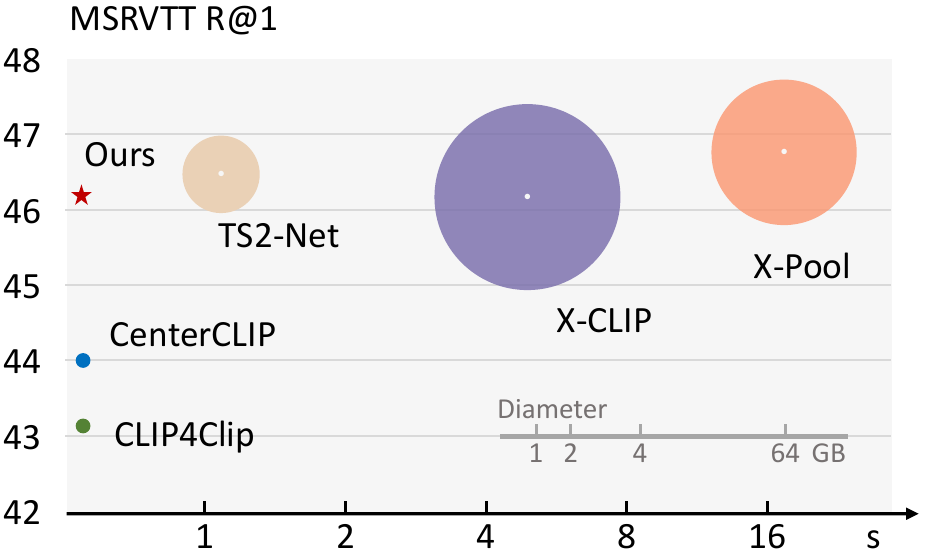}
    \caption{The performance (\ie, R@1), retrieval time, and memory usage during retrieval for baseline models and ours on the MSRVTT dataset. The center of the bubble indicates the value of R@1. The diameter of the bubble or star is proportional to the memory usage (GB) while the horizontal axis indicates the inference time (s). 
    }
    \label{fig:efficiency}
\end{figure}

Adapting CLIP to the video domain is non-trivial and requires careful consideration of both efficiency and effectiveness.
In CLIP4Clip~\cite{luo2022clip4clip}, the authors directly mean-pool the frame representations extracted by the CLIP image encoder to get the video representation and use it to calculate the cosine similarity with the text representations during retrieval.
However, the mean-pooling of frame representations may lose some essential semantic details of the video and hamper the retrieval performance.
Thus, more advanced methods, such as~\cite{gao2021clip2tv,gorti2022x,jin2022expectation,liu2022ts2,ma2022x},
generate the video representation by applying various cross-modal temporal fusion approaches on the frame representations, using text queries as the condition.
While achieving state-of-the-art results, these methods encounter severe efficiency issues in practice, as the text-conditioned fusion of each video has to be performed on-the-fly for every incoming text query.
Even with a lightweight fusion module (compared to the CLIP backbone), its computation cost grows geometrically as the number of videos and texts increases.

Formally, given a query set of $N_t$ texts with an average length of $N_w$ words and a candidate set of $N_v$ videos where each video contains $N_f$ frames.
Then, the space and time complexities are $\mathcal{O}(N_v N_t N_f)$ for the text-conditioned fusion in X-Pool~\cite{gorti2022x} and TS2-Net~\cite{liu2022ts2}, and $\mathcal{O}(N_v N_t N_f N_w)$ for that of X-CLIP~\cite{ma2022x}. 
While for CLIP4Clip, the complexity is $\mathcal{O}(N_v N_t)$ as it only requires a simple dot-product between the text and mean-pooled frame representations, although its performance is inferior to X-Pool and X-CLIP.
To better reveal this gap, we show an example in Figure~\ref{fig:efficiency} about the real-world efficiency of several methods while omitting the backbone computation.
Here, we set $N_v=16384$, $N_t=512$, $N_f=12$, and $N_w=10$.
From the figure, with large $N_t$ and $N_v$, the latency and memory consumption for text-conditioned temporal fusion methods~\cite{gorti2022x,liu2022ts2,ma2022x} are orders of magnitude higher than text-agnostic temporal fusion (\ie, mean-pooling)~\cite{luo2022clip4clip,zhao2022centerclip}, 
and can rapidly become enormous in large-scale scenarios.

On the other hand, the backbone computation of CLIP is much less of a burden in real-world retrieval systems, as the frame representations of the video can be pre-computed offline and reused for different text queries. 
Therefore, a more practical CLIP-based text-video retrieval method should focus on improving the backbone representation ability while keeping the cross-modal interaction as simple as possible.
Motivated by this, we propose a simple and efficient adaptation method for CLIP to facilitate its ability to capture both the global and detailed semantics of videos.

\begin{figure*}[t]
\centering
\includegraphics[width=0.95\linewidth]{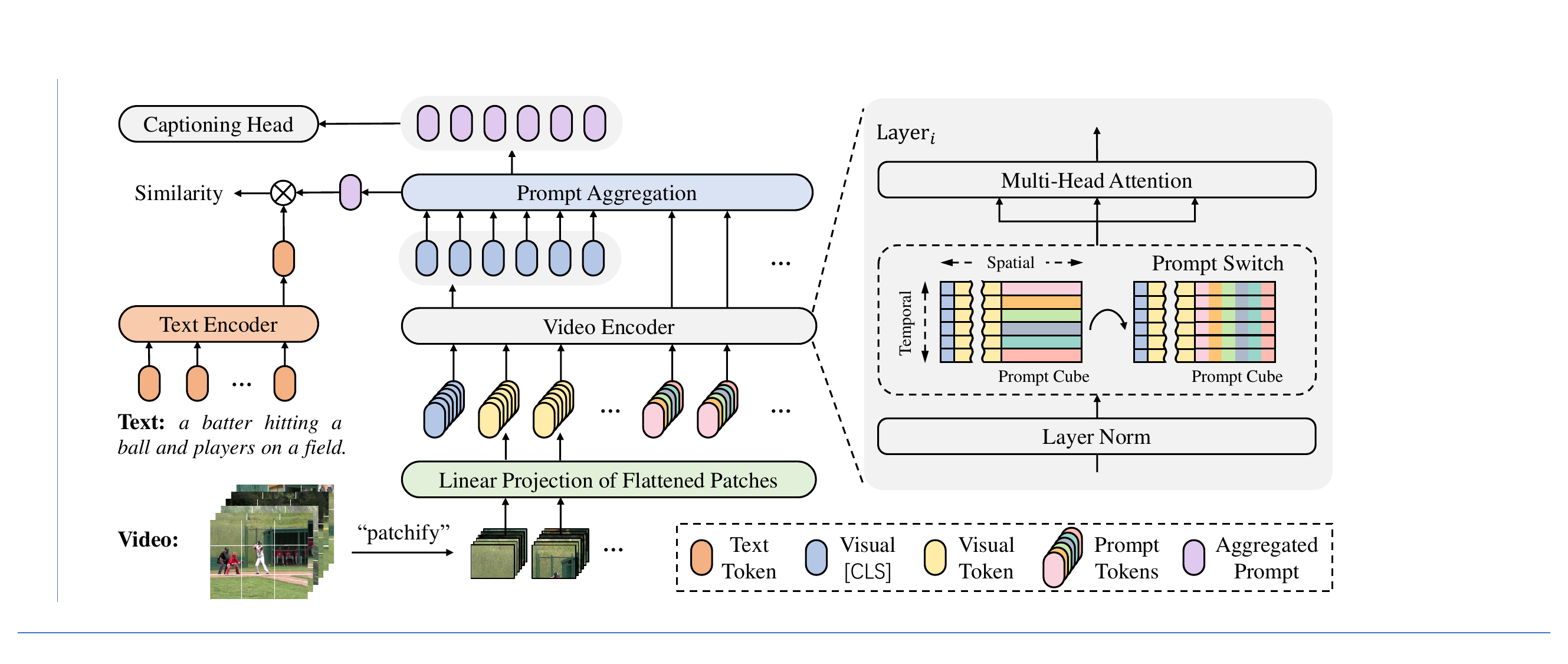}
\caption{Overall architecture. 
For a video clip with $N_f$ frames, we accompany it with $N_f\times N_f$ prompt tokens, resulting in a 3D prompt cube. 
We show the Prompt Switch operation on the right. 
For simplicity, we omit the feed-forward network and shot-cut connections in the ViT layer.
We provide more details for Prompt Aggregation in Figure~\ref{fig:captioning}.
$\otimes$ is cosine similarity.}
\label{fig:overall}
\end{figure*}

Concretely, we first feed a tiny ($\sim$0.1M) ``\textbf{Prompt Cube}'' into the image encoder of CLIP, which is a 3D tensor spanning over the spatial, temporal, and channel axis, as shown in the right of Figure~\ref{fig:overall}.\footnote{The channel axis is omitted for simplicity.}
It is designed to have the same temporal and spatial sizes and is concatenated with the patch tokens alongside the spatial axis.
To propagate temporal semantics among different frames, we \emph{switch} the spatial and temporal dimensions of the prompt cube before each self-attention layer, so that the prompt cube builds up a \emph{peer-to-peer} connection between every two-frame pair.
In this way, our modified CLIP model enjoys an improved global semantic modeling ability thanks to the comprehensive spatial-temporal modeling between the prompt cube and the patch tokens of all frames, 
while only bringing negligible extra parameters and computations.
This also allows the prompt cube to serve as a compact summarization of the whole video, 
and further enables us to design a CLIP-guided Prompt Aggregation approach and obtain the frame representations from the prompt cube.
Then, we use naive mean-pooling instead of cross-modal fusion on these frame representations to get the final video representation.

Moreover, since we will not use any fine-grained cross-modal interaction modules in our model, we adopt an Auxiliary Video Captioning objective as an alternative to provide fine-grained guidance in the semantic space when learning video representations.
Specifically, we introduce a light captioning head on top of our modified CLIP image encoder during training, which takes the frame representations aggregated from the prompt cube as input and generates the paired text of the input video.
This auxiliary objective plays a critical role because CLIP's original contrastive learning objective is relatively easy to fit due to the lack of in-batch negatives during training (which is generally the case in text-video retrieval).
During inference, the light captioning branch is removed, thus it incurs no extra computation and memory consumption.

We verify the effectiveness of the proposed method on three text-video retrieval benchmarks, \ie MSR-VTT~\cite{xu2016msr}, MSVD~\cite{chen2011collecting}, and LSMDC~\cite{rohrbach2015long}, where our method consistently achieves state-of-the-art performance while being significantly more efficient than the previous state of the arts. 
We also provide extensive performance analyses to show the superiority of our proposed method.

\section{Related Work}
Text-video retrieval is one of the fundamental tasks in video-language modeling and has lots of applications in the industry.
Here, we broadly classify the previous works into three categories and briefly review them below.

\paragraph{Off-the-Shelf Single-Modal Representations}
An early trend in text-video retrieval is to use the off-the-shelf video (\eg, I3D~\cite{carreira2017quo}) and text representations (\eg, GloVe~\cite{pennington2014glove}) and design sophisticated feature encoding methods or multi-modal fusion mechanisms on them to improve the retrieval performance~\cite{chen2020fine,dong2021dual,kiros2014unifying,yu2018joint,yu2017end}.
For example, in~\cite{chen2020fine}, a hierarchical graph reasoning approach is proposed to decompose video-text matching into global-to-local levels and disentangle texts into a hierarchical semantic graph with three levels of events, actions, and entities.
Yu~\etal~\cite{yu2018joint} propose a joint sequence fusion model for the sequential interaction of videos and texts.
Dong~\etal~\cite{dong2021dual} leverages multi-level single-modal features that represent the rich content of both video and text in a coarse-to-fine fashion.
However, their performances are limited due to the mismatched training objectives between the pre-computed representations and the text-video retrieval task.

\paragraph{Video-Language Pre-training}
To reduce the gap between pre-training and downstream tasks, large-scale pre-trained video-language representations have been proposed, such as~\cite{bain2021frozen,gabeur2020multi,ge2022bridging,lei2021less,luo2020univl,wang2022all,xu2021vlm,xu2021videoclip,xue2022advancing}.
Most of these models are pre-trained on the large-scale video-text datasets, \eg, HowTo100M~\cite{miech2019howto100m} and WebVid-2M~\cite{bain2021frozen}, which boosts promising cross-modal video understanding~\cite{li2022align,li2020hero,luo2020univl}.
One line of works~\cite{bain2021frozen,gabeur2020multi,xu2021videoclip} uses two independent encoders for video and text and then projects them into a common latent space.
These methods often adopt a contrastive loss to distinguish the paired video-text data.
For example, Bain~\etal~\cite{bain2021frozen} design an end-to-end trainable model, aiming to take advantage of both large-scale image and video captioning
datasets.
Gabeur~\etal~\cite{gabeur2020multi} propose a multi-modal transformer that can extract features at different moments and from different modalities (\eg, audio or speech) in a video.
The other line of works~\cite{lei2021less,luo2020univl,xu2021vlm} employs a single cross-modal encoder, which concatenates the video and text sequences as inputs and models them jointly in the transformer, followed by a binary classifier predicting whether these videos and texts are aligned.
Despite they can build fine-grained associations between video-text tokens, they need to input each video-text candidate pair into the model for matching score calculation during inference and thus hampers efficiency.
Besides, although the idea of video-language pre-training is promising, due to the high cost of collecting wild videos, its scale is generally much smaller than image-language pre-training, leading to an unsatisfactory generalization ability.
Thus, like~\cite{gorti2022x,luo2022clip4clip}, we seek to boost from the image-language pre-training model (\eg, CLIP~\cite{radford2021learning}) for text-video retrieval.

\paragraph{CLIP-based Adaptation}
Recently, due to the great advantage of CLIP~\cite{radford2021learning} for vision-and-language representation learning, many works~\cite{fang2023uatvr,gao2021clip2tv,jiang2022cross,jin2022expectation,ma2022x,wang2022disentangled,xue2022clip,zhang2023multimodal} seek to transfer the knowledge of CLIP to text-video retrieval tasks.
Roughly, the existing works transfer CLIP from views of feature aggregation~\cite{fang2021clip2video,gorti2022x,luo2022clip4clip,ma2022x,zhao2022centerclip}, representation alignment~\cite{fang2021clip2video,ma2022x,wang2022disentangled}, and post pre-training~\cite{wu2022cap4video,xue2022clip}.
Specifically, X-Pool~\cite{gorti2022x} designs a cross-modal attention model, seeking to enable the model to only focus on the relevant video frames conditioned on a given text.
TS2-Net~\cite{liu2022ts2} adapts CLIP by introducing a token shift module and a token selection module, which capture the temporal information and remove unimportant tokens, respectively.
X-CLIP~\cite{ma2022x} calculates both the coarse- and fine-grained similarity for higher retrieval accuracy.
While these methods can be effective, their retrieval efficiency is low due to the coupling of video and text in the cross-modal fusion process.
A more ideal way should focus on improving the representation ability of the backbone while maintaining the cross-modal interaction as efficiently as possible.
We follow this principle in our method.

\section{Method}
The overview of our method is shown in Figure~\ref{fig:overall}. 
It is based on a pre-trained CLIP with a ViT~\cite{dosovitskiy2020image}-based image encoder.
Given an input video, we first split each video frame into fix-sized non-overlapping patches and linearly project them into 1D patch embeddings.
Following CLIP, a \texttt{[CLS]} embedding is concatenated to the embedding sequence of each frame, which is pre-trained to capture the local semantics within the sequence.
Then, in the ViT encoder, our proposed ``\textbf{Prompt Cube}'' bridges the global semantic information from the patch embeddings of all frames through a Prompt Switch operation.
After that, we feed the output \texttt{[CLS]} embeddings of the video frames and the final prompt cube into a Prompt Aggregation module, where the prompt cube is aggregated into 1D vectors according to the \texttt{[CLS]} embeddings, 
and is further enhanced with fine-grained semantics through an Auxiliary Captioning Objective.
To ensure an efficient measurement of the text-video similarity, we avoid using cross-modal fusion modules and directly average these aggregated vectors into the final video representation.
A simple dot product operation is used to compute the similarity.
More details are in the following.

\subsection{Prompt Cube for Bridging Global Semantics}

At the core of our method is the proposed Prompt Cube. 
It is designed to capture the rich temporal semantics of the whole video while bringing negligible modifications and computations to the CLIP image encoder.
Formally, given an input video clip with $N_f$ frames, we first obtain its patch embeddings $\bm{V}\in \mathbb{R}^{N_f\times L \times D}$, where $L$ indicates the size of the spatial dimension, \ie, the number of patches divided from a video frame plus one \texttt{[CLS]} embedding. 
$D$ is the embedding size.
Then, our prompt cube is constructed as a 3D tensor $\bm{P} \in \mathbb{R}^{N_f\times N_f \times D}$.
At the start of the ViT layer, we concatenate $\bm{V}$ and $\bm{P}$ alongside the spatial dimension, denoted by $[\bm{V};\bm{P}] \in \mathbb{R}^{N_f \times (L + N_f) \times D}$, and then process them jointly.
The first two dimensions of $\bm{P}$ (corresponding to the temporal and spatial axis, respectively) have the same size, thus they can be transposed flexibly without altering the shape of $\bm{P}$.
This allows us to further propose an efficient Prompt Switch operation to exchange the local spatial semantics of each frame and the global temporal semantics from the whole video through the prompt cube.

\paragraph{Prompt Switch}
The proposed Prompt Switch operation can be defined in a one-line formula:
\begin{equation}
    \begin{aligned}
         \mathcal{T}:= [\bm{V};\bm{P}] \rightarrow [\bm{V};\bm{P}^{\top}].
     \end{aligned}
     \label{eq:prompt_transpose}
\end{equation}
See the right of Figure~\ref{fig:overall}, we apply this operation before every self-attention layer of the ViT encoder.
Then, self-attention is performed over the spatial dimension of $[\bm{V};\bm{P}]$, where the $i$-th row of the prompt cube (denoted by $\bm{p}_{i} \in \mathbb{R}^{1\times N_f\times D}$) and the patch embeddings of the $i$-th frame (denoted by $\bm{v}_i \in \mathbb{R}^{1\times L\times D}$) acquire information from each other.
In this way, for two consecutive ViT layers in the encoder, each element in the prompt cube (denoted by $\bm{p}_{i,j} \in \mathbb{R}^D$) is first attached to the $i$-th frame and communicates with $\bm{v}_i$,
and then switch to the $j$-th frame and communicates with $\bm{v}_j$.
By repeatedly performing this operation, the whole prompt cube builds up a peer-to-peer connection between every two-frame pair in the video clip, enabling comprehensive temporal modeling.


\begin{figure}[t]
    \centering
    \includegraphics[width=0.95\linewidth]{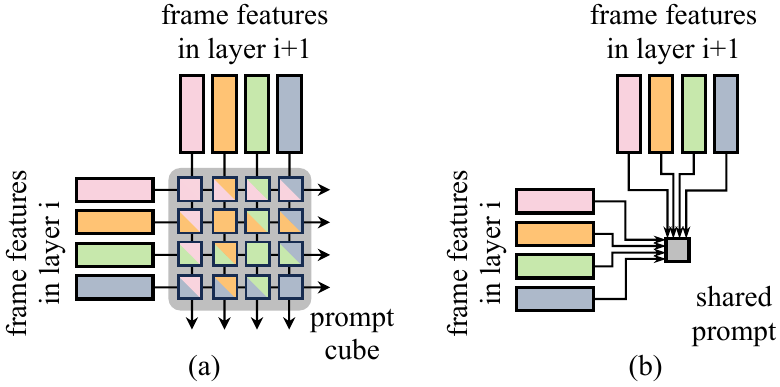}
    \caption{Video Proxy adopts a shared ``prompt'' to exchange information among all frames (b). Differently, our Prompt Switch method builds up a peer-to-peer connection between every two-frame pair (a) to obtain more comprehensive spatial-temporal modeling. This improves the representation ability of CLIP on video data. Besides, our method may also ease the learning problem as each element in the prompt cube only needs to handle the information of two frames instead of the whole video, thus improving the optimization ability of CLIP on video data.
    }
    \label{fig:peer}
\end{figure}

Some previous text-video retrieval methods have also attempted to introduce temporal adaptation to the backbone of the CLIP ViT encoder,
such as Token Shift~\cite{liu2022ts2} and Video Proxy~\cite{xue2022clip}.
Specifically, the Token Shift method shifts token embeddings from adjacent frames to the current frame, which fails to model the temporal semantics from a global perspective. 
Moreover, it damages the spatial modeling ability of the original CLIP as the information contained in the shifted tokens is no longer accessible in the current frame.
In Video Proxy, the information from all video frames is exchanged using several proxy embeddings, which lack peer-to-peer connections within the frames and thus can be inferior in the temporal modeling capacity.
We illustrate the importance of building peer-to-peer connections in Figure~\ref{fig:peer}. 
Besides, a naive full-attention approach has also been investigated where no adaptation is applied to the CLIP model except allowing its self-attention layers to attend to the patch tokens from the whole video.
However, this approach is neither effective due to the domain gap between the input data (video) and the pre-training data (image), nor efficient since the computation complexity of the self-attention layers grows quadratically \wrt the number of attended patch tokens. 
We show the superiority of our proposed prompt cube to these previous methods in Section~\ref{sec:temporal_modeling}.

\paragraph{Prompt Aggregation}
As the prompt cube acquires the global semantics through a comprehensive interaction with all patch embeddings, it can serve as a compact summarization of the video.
Therefore, we propose to obtain the final video representation from the prompt cube, instead of the \texttt{[CLS]} embedding as the original CLIP.
To achieve this, we design a CLIP-guided Prompt Aggregation module, which aggregates the output prompt cube into 1D vectors according to the final \texttt{[CLS]} embeddings of the video frames.
It is a lightweight multi-head attention (MHA) layer placed before the last layer normalization (LN) layer of the ViT encoder. 
The \texttt{[CLS]} embedding of the $i$-th frame (denoted by $\bm{c}_i \in \mathbb{R}^{1\times D}$) is used as the ``query'' while the ``key'' and ``value'' are both the prompt cube flattened over the temporal and spatial dimensions, denoted by $\hat{\bm{P}} \in \mathbb{R}^{N_f^2\times D}$.
Let $\bar{\bm{p}_i} \in \mathbb{R}^{1\times D}$ be the aggregated prompt vector for the $i$-th frame, we have
\begin{equation}
\begin{split}
\bar{\bm{p}_i} := \text{LN}(\bm{c}_i + \text{MHA}(\bm{c}_i, \hat{\bm{P}}, \hat{\bm{P}})), ~~~~\forall i \in [1, N_f],
\end{split}
\end{equation}
where LN denotes the last layer normalization layer of the CLIP ViT encoder.
Notably, the last linear projection weight in the MHA layer is initialized as a $0$ tensor to ensure that $\bar{\bm{p}_i}$ is optimized from the original output of the CLIP image encoder.
The final video representation is the naive mean-pooling of the normalized prompt vectors, \ie,
\begin{equation}\label{eq:meanpool}
\bm{x}:=\frac{1}{N_f}\sum_{i=1}^{N_f} \frac{\bar{\bm{p}_i}}{\|\bar{\bm{p}_i}\|}.
\end{equation}
We provide a detailed illustration for the Prompt Aggregation module on the left of Figure~\ref{fig:captioning}. 
To enhance the learning of temporal information for the prompt cube, inspired by~\cite{zhu2017flow}, we adopt a frame sampling strategy where we randomly sample $k<N_f$ prompt vectors for the final mean-pooling operation in Eq.~(\ref{eq:meanpool}) during training.

\subsection{Learning Detailed Semantics via Captioning}
As we have discussed in Section~\ref{sec:introduction}, the key criterion in the design of our model architecture is to ensure the model contains no extra cross-modal interaction procedure during inference except computing the cosine similarity between text and video representations.
While this ensures the efficiency of the similarity measurement in large-scale production systems, it also prevents the video representation from utilizing the fine-grained semantic information from the text query.
To aid this, inspired by~\cite{chen2022learning,yu2022coca}, we propose to learn our video representation with an Auxiliary Captioning objective, which alternatively provides fine-grained guidance in the semantic space during training.

\begin{figure}[t]
    \centering
    \includegraphics[width=0.90\linewidth]{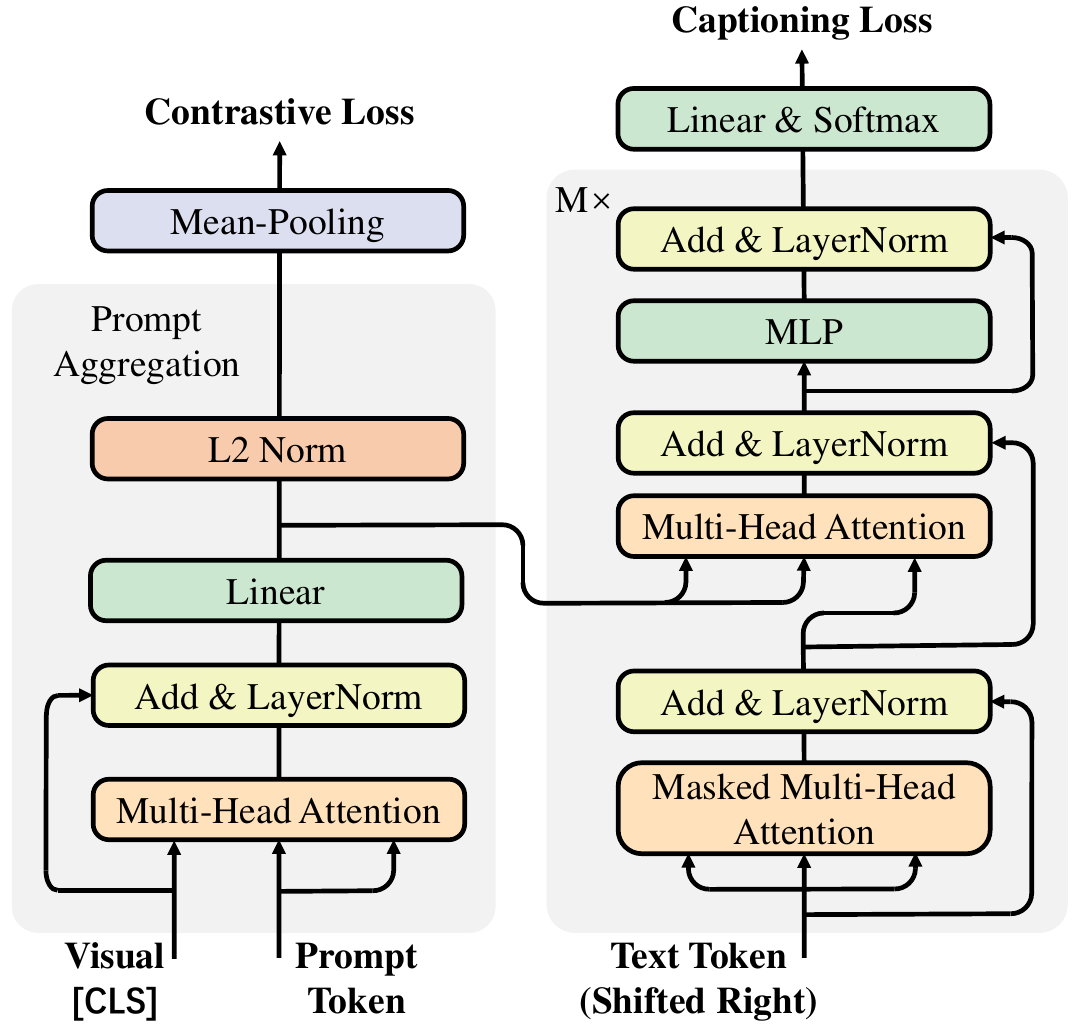}
    \caption{Detailed architecture of Prompt Aggregation (left) and Auxiliary Captioning Head (right).
    }
    \label{fig:captioning}
\end{figure}

\paragraph{Auxiliary Captioning Head}
As shown on the right of Figure~\ref{fig:captioning}, the Auxiliary Captioning Head, denoted by $\mathcal{H}$, is a stack of $M$ transformer decoders.
We first shift the text token by one step to the right, and feed it into a Masked MHA layer, where each text token only attends to its preceding tokens.
Then, another MHA layer is used, where the text tokens attend to the aggregated prompt vectors $\{\bar{\bm{p}_i}\}_{i=1}^{N_f}$.
The model is trained via the autoregressive Teacher Forcing~\cite{williams1989learning} scheme, \ie, the prediction in each step should maximize the likelihood of the token in the next step:
\begin{equation}
    \begin{aligned}
         \mathcal{L}_{cap} = \sum^{N_w}_{l=1} -\log p(w_l|w_{<l}, \{\bar{\bm{p}_i}\}_{i=1}^{N_f}),
     \end{aligned}
     \label{eq:captioning_AT}
\end{equation}
where $w_l$ indicates the $l$-th token in the text, and $w_{<l}$ denotes the tokens before $w_l$.
$N_w$ is the total number of tokens.
Notably, we discard some commonly occurred words (\eg, ``a'', ``an'' and ``the'') by performing a Term Frequency Inverse Document Frequency (TF-IDF) weighting for each word like~\cite{vedantam2015cider}, which ensures that the captioning model focuses only on the informative words. 
During inference, $\mathcal{H}$ is dropped, thus no extra computation is incurred.

\paragraph{Overall Training Objective}
Our overall training objective consists of a contrastive loss and a captioning loss. 
Typically, the contrastive loss is defined as 
\begin{equation}
\begin{aligned}
 & \mathcal{L}_{v2t} = - \frac{1}{B}\sum_{i=1}^B \log \frac{\exp(\bm{x}_i^{\top}\bm{y}_i/\tau)}{\sum_{j=1}^B\exp(\bm{x}_i^{\top}\bm{y}_j/\tau)}, \\
 & \mathcal{L}_{t2v} = - \frac{1}{B}\sum_{i=1}^B \log \frac{\exp(\bm{y}_i^{\top}\bm{x}_i/\tau)}{\sum_{j=1}^B\exp(\bm{y}_i^{\top}\bm{x}_j/\tau)},\\
 & \mathcal{L}_{con} = \frac{1}{2}(\mathcal{L}_{v2t} + \mathcal{L}_{t2v}),
\end{aligned}
\label{eq:info_nce_loss}
\end{equation}
where $B$ is the size of the mini-batch, $\tau$ is a learnable temperature parameter.
$\bm{x}_i$ and $\bm{y}_i$ are the video representation (obtained by Eq.~(\ref{eq:meanpool}) and text representation of the $i$-th video-text pair in the mini-batch.
The final loss function is
\begin{equation}
\begin{aligned}
 \mathcal{L} = \mathcal{L}_{con} + \lambda\mathcal{L}_{cap},
\end{aligned}
\label{eq:final_loss}
\end{equation}
where $\lambda$ is the weighting hyper-parameter.

\section{Experiment}

\paragraph{Dataset and Metrics}
We evaluate the effectiveness of our method on three widely used text-video retrieval datasets, including MSR-VTT~\cite{xu2016msr}, MSVD~\cite{chen2011collecting} and LSMDC~\cite{rohrbach2015long}.
\textbf{MSR-VTT}~\cite{xu2016msr} consists of 10,000 videos with 200K descriptions. 
We follow the setting in previous works~\cite{gabeur2020multi,yu2018joint},  training our models with 9,000 videos and evaluating them on the 1K-A test set.
\textbf{MSVD}~\cite{chen2011collecting} contains $1,970$ videos with about 120K captions, where the train, validation, and test splits contain $1,200$, $100$ and $670$ videos, respectively.
\textbf{LSMDC}~\cite{rohrbach2015long} has $118,081$ video-caption pairs, where $109,673$ videos are used for training, $7,408$ videos for validation, and $1,000$ videos for testing. 
Following~\cite{gorti2022x,liu2022ts2,luo2022clip4clip}, we report the results of Recall@$K$ (R@$K$, $K=1, 5, 10$), and Mean Rank (MnR) for quantitative evaluation. 
Besides, we sum the value of R@$1$, R@$5$, and R@$10$ for both text-video and video-text retrieval tasks as an overall evaluation metric, named Meta Sum.

\begin{table}[t]
\centering
\resizebox{1.0\linewidth}{!}{
\begin{tabular}{l|lccc}
\toprule
Method & R@1$\uparrow$    & R@5$\uparrow$    & R@10$\uparrow$   & MnR$\downarrow$ \\
\midrule
baseline CLIP & 43.1 & 70.9 & 80.4 & 16.0 \\
+~Prompt Switch  & 44.8 \textcolor{ForestGreen}{(+1.7)}& 71.9 & 80.9 & 15.2 \\
\textcolor{gray}{~~~~+~Temporal Transformer} & \textcolor{gray}{44.2} & \textcolor{gray}{71.4} & \textcolor{gray}{81.1} & \textcolor{gray}{15.6} \\
~~~~+~Prompt Aggregation  & 45.4 \textcolor{ForestGreen}{(+0.6)} & 71.9 & 81.1 & 14.9 \\
~~~~~~~~+~Captioning Loss & \textbf{46.1} \textcolor{ForestGreen}{(+0.7)} & \textbf{72.8} & \textbf{81.8} & \textbf{14.4}  \\
\bottomrule
\end{tabular}%
}
\caption{Performance analysis of our model components for text-video retrieval on MSRVTT 1K-A test set.
Both Prompt Aggregation and Temporal Transformer are applied on top of our Prompt Switch.
The Captioning Loss is applied only on top of the Prompt Aggregation.
}
\label{tab:ablation_ours}%
\end{table}%

\begin{table}[t]
\centering
\resizebox{1.0\linewidth}{!}
{
\begin{tabular}{l|cccc}
\toprule
Method & R@1 $\uparrow$   & R@5 $\uparrow$   & R@10 $\uparrow$  & MnR $\downarrow$  \\
\midrule
Temporal Transformer & 44.2 & 71.4 & 81.1 & 15.6 \\
Token Shift~\cite{liu2022ts2} & 43.2 & 70.7 & 79.8 & 15.9 \\
Video Proxy~\cite{xue2022clip} & 45.2 & 71.0 & \textbf{81.5} & 15.3\\
Full Attention & 43.8 & 71.4 & 80.9 & 16.4 \\ \midrule
Prompt Switch \& Aggregation & \textbf{45.4} & \textbf{71.9} & 81.1 & \textbf{14.9} \\
\bottomrule
\end{tabular}%
}
\caption{Comparisons with other temporal modeling methods for text-video retrieval on MSRVTT 1K-A test set. 
}
\label{tab:ablation_backbone}%
\end{table}%

\begin{table}[t]
\centering
\resizebox{1.0\linewidth}{!}
{
\begin{tabular}{l|cccl}
\toprule
Method & R@1 $\uparrow$   & R@5 $\uparrow$   & R@10 $\uparrow$  & Mem.~/~Time \\
\midrule
\multicolumn{5}{c}{Text $\Rightarrow$ Video} \\
\midrule
TS2-Net*~\cite{liu2022ts2} & 46.5  & 73.6  & \textbf{83.3}  & 0.3G~/~0.2s \\
X-Pool~\cite{gorti2022x} & 46.9  & 72.8  & 82.2  & 3.9G~/~3.9s \\
\midrule
ours (mean pool)  & 46.1 & 72.8 & 81.8 & \textbf{2.0}M~/~\textbf{8.1}ms \\
ours (attention pool) & 46.4 & 72.9 & 82.4 & 0.2G~/~0.2s \\
ours (top-$3$ pool)  & 46.7 & 73.4 & 82.0 & 0.3G~/~0.3s \\ 
ours (X-Pool~\cite{gorti2022x}) & \textbf{47.8} & \textbf{73.9} & 82.2 & 3.9G~/~3.9s \\
\midrule
\multicolumn{5}{c}{Video $\Rightarrow$ Text} \\
\midrule
TS2-Net*~\cite{liu2022ts2} & 44.5  & 73.8  & 83.2  & 0.3G~/~0.2s \\
X-Pool~\cite{gorti2022x} & 44.4  & 73.3  & 84.0   & 3.9G~/~3.9s \\
\midrule
ours (mean pool)  & 44.8 & 73.7 & 82.4 & \textbf{2.0}M~/~\textbf{8.1}ms \\
ours (attention pool) &  45.4 & 73.9 & 83.2 & 0.2G~/~0.2s \\
ours (top-$3$ pool) & 45.2 & 73.6 & 83.7 & 0.3G~/~0.3s \\
ours (X-Pool~\cite{gorti2022x}) &  \textbf{46.0} & \textbf{74.3} & \textbf{84.8} & 3.9G~/~3.9s \\
\bottomrule
\end{tabular}%
}
\caption{Performance and complexity comparisons with different temporal fusion methods on MSRVTT 1K-A test set.
* denotes the results reproduced by official code \& setting.
}
\label{tab:ablation_pooling}%
\end{table}%

\paragraph{Implementation Details}
Following previous work~\cite{gorti2022x,luo2022clip4clip}, both the text and video encoders are initialized with the pre-trained CLIP (ViT-B/32)~\cite{radford2021learning}.
The prompt cube is randomly initialized from a Gaussian distribution with a
zero mean and 0.02 std.
During training, we uniformly sample 6 frames from each video and resize all video frames into $224\times 224$. 
Therefore, the size of the spatial and temporal dimensions of the prompt cube is set to 6 by default.
While for the testing, 12 frames are used as in previous works, so we split them into two 6-frame chunks through interval sampling to make the temporal dimension compatible with the prompt cube.
The Prompt Aggregation is applied on the prompt cubes of all chunks.
The number of decoder layers in the captioning head is $3$. 
The hyper-parameter $k$ used for frame sampling in Eq.~(\ref{eq:meanpool}) is set to $3$.
We set the hyper-parameter $\lambda=0.5$ in Eq.~(\ref{eq:final_loss}).
The training epochs are 10 for all the datasets with a batch size of 128.
We use AdamW optimizer~\cite{loshchilov2017decoupled} with a learning rate of 3e-5 and adopt a cosine decay strategy for the learning rate.

\begin{table*}[t!]
  \centering
  \resizebox{0.9\linewidth}{!}
  {
    \begin{tabular}{l|cccc|cccc|c}
    \toprule
    \multirow{2}[2]{*}{Methods} & \multicolumn{4}{c|}{Text $\Rightarrow$ Video} & \multicolumn{4}{c|}{Video $\Rightarrow$ Text} & \multirow{2}[2]{*}{Meta Sum $\uparrow$} \\
          & R@1 $\uparrow$   & R@5 $\uparrow$   & R@10 $\uparrow$  & MnR $\downarrow$   & R@1    & R@5$\uparrow$    & R@10$\uparrow$  & MnR $\downarrow$  \\
    \midrule
    \textit{cross-modal temporal fusion} \\
    CLIP2TV~\cite{gao2021clip2tv}   &  46.1 & 72.5  & 82.9 & 15.2  & 43.9  & 73.0  & 82.8 & 11.1  & 401.2  \\
    CLIP2Video~\cite{fang2021clip2video}   &  45.6 &  72.6  & 81.7 & 14.6  & 43.3  & 72.3  & 82.1 & 10.2  & 397.6   \\
    TS2-Net*~\cite{liu2022ts2} & 46.5  & 73.6  & 83.3  & 13.9  & 44.5  & 73.8  & 83.2  & 9.2 & 404.9  \\
    EMCL~\cite{jin2022expectation}  & 46.8  & 73.1  & 83.1  & - & 46.5  & 73.5  & 83.5  & -  & 406.5  \\
    X-CLIP~\cite{ma2022x} & 46.1  & 73.0    & 83.1  & 13.2  & 46.8  & 73.3  & 84.0  & 9.1 & 406.3  \\
    DRL~\cite{wang2022disentangled}   & 47.4  & 74.6  & 83.8  & - & 45.3  & 73.9  & 83.3  & - & 408.3 \\
    X-Pool~\cite{gorti2022x} & 46.9  & 72.8  & 82.2  & 14.3  & 44.4  & 73.3  & 84.0   & 9.0 & 403.6   \\
    ours (X-Pool) & \textbf{47.8} & \textbf{73.9} & 82.2 & 14.1 &  46.0 & \textbf{74.3} & \textbf{84.8} & \textbf{8.5} & \textbf{409.0} \\
    \midrule
    \textit{text-agnostic temporal pooling} \\
    CLIP4Clip$^\dagger$ (seqTransf)~\cite{luo2022clip4clip} & 44.5  & 71.4  & 81.6  & 15.3  & 42.7  & 70.9  & 80.6   & 11.6 & 391.7  \\
    CenterCLIP$^\dagger$ (spectral)~\cite{zhao2022centerclip} & 44.2  & 71.6  & 82.1  & 15.1  & 42.8  & 71.7  & 82.2   & 11.1 & 394.6  \\
    X-CLIP (mean pool)~\cite{ma2022x} & 43.0    & 70.7  & 81.6   & 16.3  & 43.0    & 70.2  & 81.2  & 11.5 & 389.7  \\
    TS2-Net* (mean pool)~\cite{liu2022ts2} & 44.4  & 72.1 & \textbf{82.2} & 14.6 & 43.7 & 70.8 & 80.4 & 11.6 & 393.6  \\
    ours (mean pool) &   \textbf{46.1}   &  \textbf{72.8}    &  81.8  &  \textbf{14.4}    &  \textbf{44.8}  &  \textbf{73.7}  &  \textbf{82.4}  &  \textbf{9.9} & \textbf{401.6} \\
    \bottomrule
    \end{tabular}%
    }
  \caption{Comparisons with state-of-the-arts on MSRVTT. $^\dagger$ both CLIP4Clip and CenterCLIP have multiple versions in their papers, here we choose the versions with the highest Meta Sum.  * denotes the results reproduced by official code \& setting.}
  \label{tab:msrvtt}%
\end{table*}%

\begin{table*}[t]
  \centering
  \resizebox{0.9\linewidth}{!}
  {
    \begin{tabular}{l|cccc|cccc|c}
    \toprule
    \multirow{2}[2]{*}{Methods} & \multicolumn{4}{c|}{Text $\Rightarrow$ Video} & \multicolumn{4}{c|}{Video $\Rightarrow$ Text} & \multirow{2}[2]{*}{Meta Sum $\uparrow$}\\
          & R@1 $\uparrow$   & R@5 $\uparrow$   & R@10 $\uparrow$  & MnR $\downarrow$   & R@1    & R@5$\uparrow$    & R@10$\uparrow$  & MnR $\downarrow$  \\
    \midrule
    \textit{cross-modal temporal fusion} \\
    CLIP2TV~\cite{gao2021clip2tv}  & 47.0  & 76.5  & 85.1 & 10.1  &  - &  - & - & -  & - \\
    CLIP2Video~\cite{fang2021clip2video}   &  47.0 & 76.8  & 85.9 & 9.6  & 58.7  & 85.6  &  91.6 & 4.3  & 445.6 \\
    X-CLIP~\cite{ma2022x} & 47.1  &  77.8  & -  & 9.5  & 60.9  & 87.8 & -  & 4.7 & -\\
    X-Pool~\cite{gorti2022x} &  47.2 &  77.4  & 86.0  & 9.3  & 66.4  & 90.0 & 94.2  & 3.3 & 461.2\\
    \midrule
    \textit{text-agnostic temporal pooling} \\
    CLIP4Clip$^\dagger$  (seqTransf)~\cite{luo2022clip4clip} &  45.2 &  75.5  & 84.3  & 10.3  & 62.0  & 87.3 & 92.6  & 4.3 & 446.9\\
    CenterCLIP$^\dagger$  (spectral)~\cite{zhao2022centerclip} & \textbf{47.4}  & 76.5 & 85.2 & 9.7  & 62.7  & 88.1 & 92.8  & 4.1 & 452.7\\
    ours (mean pool) &  47.1 & \textbf{76.9} & \textbf{86.1}  & \textbf{9.5}  & \textbf{68.5}  & \textbf{91.8} & \textbf{95.6}  & \textbf{2.8} & \textbf{466.0}\\
    \bottomrule
    \end{tabular}%
    }
  \caption{Comparisons with state-of-the-arts on MSVD. $^\dagger$ both CLIP4Clip and CenterCLIP have multiple versions in their papers, here we choose the versions with the highest Meta Sum values.}
  \label{tab:msvd}%
\end{table*}%

\subsection{Performance Analysis}
We first thoroughly analyze the model designs and critical components of our proposed method and then verify their effectiveness.
The experiments are conducted on the MSRVTT dataset and evaluated on the 1K-A test set. 

\vspace{-5pt}
\paragraph{Model Components}
As shown in Table~\ref{tab:ablation_ours}, we conduct an ablation study on the Prompt Switch, Prompt Aggregation, and Auxiliary Captioning Objective by introducing them gradually into the model for text-video retrieval. 
The baseline CLIP model directly uses the mean-pooled \texttt{[CLS]} embeddings of the video frames as the final video representation, and is trained using only contrastive loss.
Specifically, when incorporating the Prompt Switch mechanism, the result of R@1 increases significantly (\ie, from $43.1$ to $44.8$) compared with the baseline CLIP, which demonstrates the effectiveness of the proposed Prompt Switch.
With the help of Prompt Aggregation, it further improves $0.6$ and achieves $45.4$ in R@1.
Our final model containing the Auxiliary Captioning Loss further improves the performance in R@1, R@5, R@10, and MnR by a large margin.
These results demonstrate that all the proposed components contribute clearly to the final performance.
Besides, we also compare the performance of an \textbf{alternative} temporal aggregation method to Prompt Aggregation termed Temporal Transformer, which adopts an additional transformer layer on the \texttt{[CLS]} embeddings of all video frames and averages the output.
From Table~\ref{tab:ablation_ours}, Prompt Aggregation is better than Temporal Transformer on all metrics.

\begin{table*}[t]
  \centering
  \resizebox{0.9\linewidth}{!}
  {
    \begin{tabular}{l|cccc|cccc|c}
    \toprule
    \multirow{2}[2]{*}{Methods} & \multicolumn{4}{c|}{Text $\Rightarrow$ Video} & \multicolumn{4}{c|}{Video $\Rightarrow$ Text} & \multirow{2}[2]{*}{Meta Sum $\uparrow$} \\
     & R@1$\uparrow$    & R@5$\uparrow$  & R@10$\uparrow$ & MnR$\downarrow$ & R@1$\uparrow$    & R@5$\uparrow$  & R@10$\uparrow$ & MnR$\downarrow$\\
    \midrule
    \textit{cross-modal temporal fusion} \\
    X-Pool~\cite{gorti2022x} & 25.2 & 43.7 & 53.5 & 53.2 & 22.7 & 42.6 & 51.2 & 47.4 & 238.9 \\
    TS2-Net~\cite{liu2022ts2} & 23.4 & 42.3 & 50.9 & 56.9 & - & - & - & - & - \\
    EMCL~\cite{jin2022expectation} & 23.9 & 42.4 & 50.9 & - & 22.2 & 40.6 & 49.2 & - & 229.2\\
    X-CLIP~\cite{ma2022x} & 23.3 & 43.0 & - & 56.0 & 22.5 & 42.2 & - & 50.7 & -\\
    \midrule
    \textit{text-agnostic temporal pooling} \\
    CLIP4Clip$^\dagger$ (seqLSTM)~\cite{luo2022clip4clip} & 21.6 & 41.8 & 49.8 & 58.0 & 20.9 & 40.7 & 49.1 & 53.9 & 223.9 \\
    CenterCLIP$^\dagger$ (k-medoids++)~\cite{zhao2022centerclip} & 21.9 & 41.1 & \textbf{50.7} & \textbf{55.6} & 21.1 & 41.2 & 50.2 & \textbf{48.7} & 226.2\\
    ours (mean pool) & \textbf{23.1} & \textbf{41.7} & {50.5} & 56.8 & \textbf{22.0} & 40.8 & \textbf{50.3} & 51.0 & \textbf{228.4}\\
\bottomrule
    \end{tabular}%
    }
  \caption{Comparisons with state-of-the-arts on LSMDC. $^\dagger$ both CLIP4Clip and CenterCLIP have multiple versions in their papers and here we choose the versions with the highest Meta Sum values.}
  \vspace{-3pt}
  \label{tab:lsmdc}%
\end{table*}%

\vspace{-5pt}
\paragraph{Comparison on Temporal Modeling Methods} \label{sec:temporal_modeling}
To investigate the effectiveness of our temporal modeling method, \ie, Prompt Switch $+$ Prompt Aggregation, we compare it with other four temporal modeling methods, including the Temporal Transformer, Full Attention (attention over patch tokens from all video frames), Token Shift~\cite{liu2022ts2}, and Video Proxy~\cite{xue2022clip}.
All methods are implemented on the baseline CLIP model without extra components like the Token Selection Transformer in \cite{liu2022ts2}, and are trained using our default setting.
From Table~\ref{tab:ablation_backbone}, 
our Prompt Switch \& Aggregation approach outperforms all the baselines on R@1, R@5, and MnR while achieving the second-best result on R@10, 
which demonstrates its superiority in learning global video semantics across frames.

\vspace{-5pt}
\paragraph{Cross-Modal Temporal Fusion \vs Naive Mean Pooling}
While our method is able to achieve significantly better performance than the baseline CLIP using the mean-pooling setting, we further investigate whether it is compatible with the advanced cross-modal temporal fusion methods, namely, attention pooling, top-$3$ pooling, and X-Pool~\cite{gorti2022x}.
We simply replace the final mean-pooling in Eq.~(\ref{eq:meanpool}) with these cross-modal approaches and show the performance in Table~\ref{tab:ablation_pooling}.
From the table, when adopting these cross-modal fusion methods, our models obtain clearly boosted performance, especially for X-Pool, which obtains the best performance for both text-video and video-text retrieval and outperforms the original X-Pool by a large margin. 
Moreover, even compared with the state-of-the-art cross-modal fusion methods, our model with naive mean-pooling is still competitive for both text-video and video-text retrieval.

We further measure the memory usage and latency of the compared methods during inference on the test set. 
To make the experiment setting close to real-world scenarios and for fair comparisons, we use the same pre-computed frame and text representations while only monitoring space and time consumption for the ranking procedure.
As shown in the last column of Table~\ref{tab:ablation_pooling}, our model with mean-pooling is orders of magnitude more efficient than those baseline models with cross-modal temporal fusion. 
This conclusion is generalizable to other datasets, \ie, the same model will always have the same ranking efficiency if using the same inference and evaluation settings.
This reveals the importance of using text-agnostic temporal fusion (\eg, mean-pooling) for real-world text-video retrieval.

\subsection{Comparison with the State-of-the-arts}
In this section, we compare our proposed method with state-of-the-art methods on MSRVTT, MSVD, and LSMDC dataset.
To better reveal the performance gap among the compared methods, we do not consider post-processing techniques like QB-NORM~\cite{bogolin2022cross}.
The results of both text-video and video-text retrieval tasks are presented in Tables~\ref{tab:msrvtt}, \ref{tab:msvd} and~\ref{tab:lsmdc}.
From the tables, when compared under the text-agnostic temporal fusion setting,
our model outperforms the baseline models on most of the evaluation metrics, especially for Meta Sum, where it achieves the best performance on all three datasets.
Specifically, on MSRVTT, the Meta Sum of our model is 7 points better than the second-best model; on MSVD, our method outperforms CenterCLIP~\cite{zhao2022centerclip} by 13.3; On LSMDC, our model consistently achieves the highest Meta Sum.
Moreover, when compared with methods using cross-modal temporal fusion, our model with mean-pooling is still competitive in terms of Meta Sum, while being much more efficient in practice (refer to Figure~\ref{fig:efficiency} and Table~\ref{tab:ablation_pooling} for detailed discussions).
Specifically, on MSRVTT, it outperforms CLIP2TV~\cite{gao2021clip2tv} and CLIP2Video~\cite{fang2021clip2video}; on MSVD, it surpasses all compared baselines and improves over the second-best model (X-Pool~\cite{gorti2022x}) by 4.8; on LSMDC, it is still comparable with EMCL~\cite{jin2022expectation}.
These results show that our proposed method has a better trade-off between performance and efficiency, and is more suitable for large-scale production systems.

\begin{figure}[t]
    \centering
    \subfloat[R@1 performance with different number of decoder layers]{\includegraphics[width=1.0\linewidth]{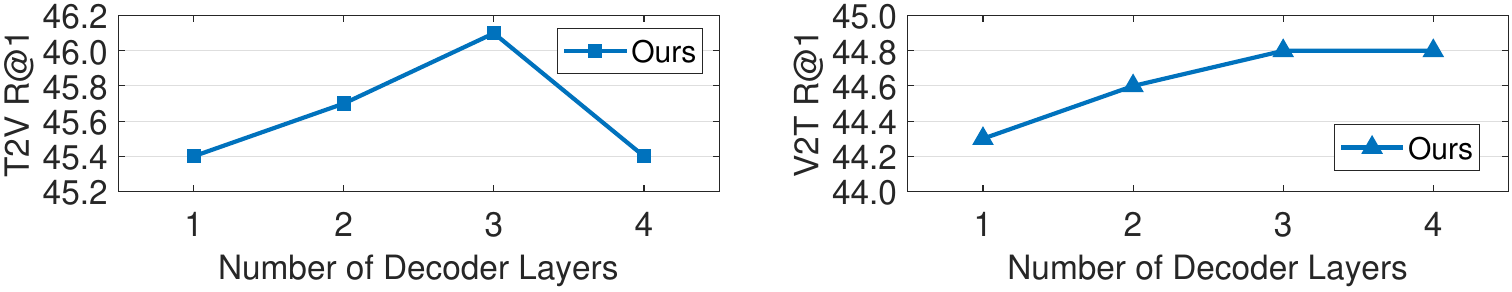}} \\
    \subfloat[R@1 performance with different number of subsampling frames]{\includegraphics[width=1.0\linewidth]{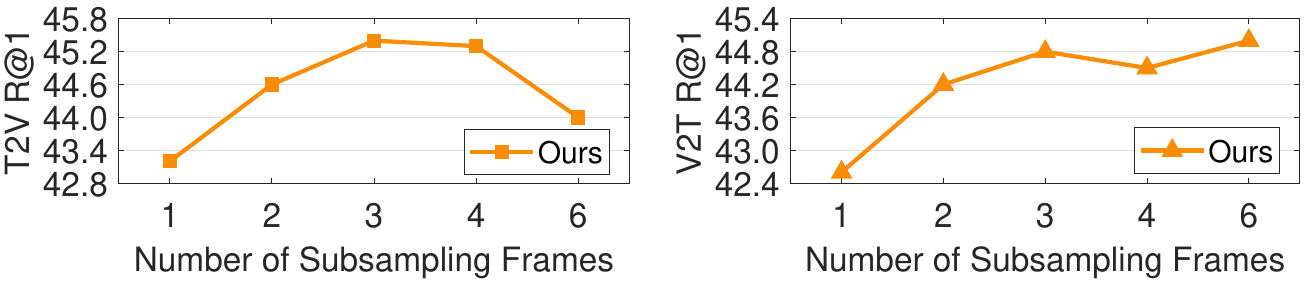}} \\
    \subfloat[R@1 performance with different value of hyper-parameter $\lambda$]{\includegraphics[width=1.0\linewidth]{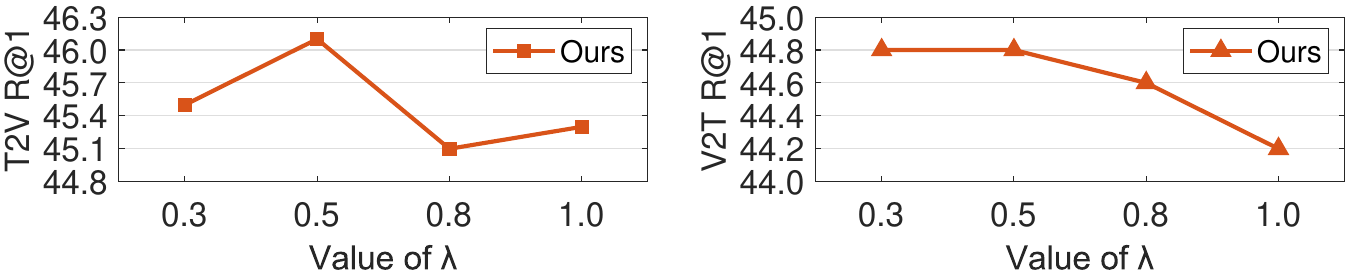}}
    \caption{We show the effects of (a) the number of captioning decoder layers; (b) the number of subsampling frames $k$ in training; (c) the values of hyper-parameter $\lambda$ in Eq.~(\ref{eq:final_loss}).} 
    \label{fig:num_layer_subsampling_lambda}
\end{figure}

\subsection{Further Discussions}

\paragraph{Number of Decoder Layers}
Generally, more layers in the captioning decoder would improve its capacity for learning fine-grained knowledge, which, however, also makes it easier to over-fit. To find a better trade-off, we study the effect of the number of layers. As shown in Figure~\ref{fig:num_layer_subsampling_lambda}(a), the performance of our model improves and reaches the peak with $3$ layers of decoder for both text-video and video-text retrieval tasks. Thus, we set the number of decoder layers to $3$ for all the experiments.

\paragraph{Number of Subsampling Frames}
In practice, we set the number of training frames as $6$ and validation frames as $12$. Then, we randomly sample $k$ frames before mean-pooling for training while using all $12$ frames for evaluation. Notably, if $k=6$, that means we average all the training frames in the mean-pooling operation.
In Figure~\ref{fig:num_layer_subsampling_lambda}(b), we observe that the performance improves when reducing the value of $k$ and achieves the peak when $k=3$, which demonstrates its effectiveness.
While further decreasing the value of $k$, the mean-pooling operation may lose a great deal of semantics derived from frames, which causes performance degradation. Therefore, we set $k=3$ for all the experiments.

\paragraph{Hyper-parameter $\lambda$ in Eq.~(\ref{eq:final_loss})}
From Figure~\ref{fig:num_layer_subsampling_lambda}(c), in the case of a small $\lambda$ (\eg, $\lambda=0.3$), the model can only achieve suboptimal performance due to insufficient exploitation of the detailed semantics. When increasing the value of $\lambda$, the performance of our model peaks at the $\lambda=0.5$ and degrades thereafter. Considering the trade-off between typical contrastive loss and captioning loss, we choose the weighting parameter $\lambda$ as $0.5$ for all the datasets.

\section{Conclusion}

In this paper, we tackle the task of text-video retrieval, where we aim to learn semantically-enhanced representations purely from the video, allowing for offline computation and reuse for different text queries.
Our method introduces a new Prompt Cube into the CLIP image encoder, which is iteratively transposed within the encoder layers to incorporate global video semantics into frame representations. We also adopt an auxiliary video captioning objective to optimize the frame representations, providing detailed guidance in the semantic space. With mean pooling fusion on the enhanced frame representations, the proposed model achieves SoTA performance on three benchmark datasets. Comprehensive experiments verify the effectiveness of all critical components of our proposed method.


{\small
\bibliographystyle{ieee_fullname}
\bibliography{egbib}
}

\end{document}